# Learning Entropy Signature for Image Representation and Classification

Jan Glaser
*Dpt. of Applied Mathematics*
*Czech Technical University in Prague*
Prague, Czech Republic
glaseja1@cvut.cz

Ivo Bukovsky
*Dpt. of Computer Science*
*University of South Bohemia in Ceske Budejovice*
Ceske Budejovice, Czech Republic
ibuk@prf.jcu.cz

Noriyasu Homma
*Dpt. of Radiological Imaging and Informatics*
*Tohoku University Graduate School of Medicine*
Sendai, Japan
homma@tohoku.ac.jp

Marcel Jirina
*Dpt.of Applied Mathematics*
*Czech Technical University in Prague*
Prague, Czech Republic
marcel.Jirina@fit.cvut.cz

***Abstract*— Learning Entropy (LE) has recently been extended to image analysis through Spatial Learning Entropy Maps (SLEMs), which are two-dimensional LE distributions that highlight unusually high learning activity across an image. Unlike conventional image descriptors, SLEMs are generated by incremental, sample-wise learning of a pretrained feedforward MLP network, where local pixel neighborhoods are presented sequentially in a fixed spatial order to predict the corresponding central pixels. Consequently, the learning activity at each image location depends not only on its local structure but also on the knowledge acquired from previously processed locations. This paper introduces Learning Entropy Signatures (LES), an image descriptor derived from SLEM using the K largest LE locations. LES captures the spatial organization of learning-relevant image structures and provides a compact representation of image content based on learning weight behavior. Experimental evaluation on image classification tasks shows that a relatively small number of K largest LE locations preserve substantial discriminative information. The results indicate a close relationship between the learning of neural weights and information relevance, extending the role of Learning Entropy from time series to images and, within images, from structural point extraction to compact image representation and classification.**

## I. Introduction

The extraction of compact and informative image representations remains a fundamental problem in image analysis, pattern recognition, and machine learning. Classical approaches address this problem either through the detection of salient image features, such as corners, junctions, and intensity discontinuities, represented by methods including Harris [1], Shi–Tomasi [2], SURF [3], and SIFT [4], or through compact object representations based on Fourier descriptors [5], moment invariants [6], shape contexts [7], and inner-distance shape descriptors [8]. More recently, deep neural networks have become a dominant source of image representations by automatically constructing feature spaces optimized for classification tasks [9], [10]. Despite substantial advances in representation learning, the underlying question remains largely unchanged: which image observations contribute most significantly to the acquisition of information by a learning system?

In several research areas, the significance of an observation is no longer defined solely by its observable properties but rather by its influence on the internal state of a learning system. Examples include Bayesian surprise [11], information-gain-based learning [12], active learning [13], intrinsic motivation systems [14], and curiosity-driven learning [15]. Although differing in formulation, these approaches share a common principle: informative observations are those that induce substantial modifications of the internal model of the learning agent. This perspective naturally motivates an learning-oriented interpretation of information relevance.

Learning Entropy (LE) was introduced as a quantitative measure of learning activity in adaptive systems [16], [17], [18]. Unlike conventional signal descriptors, LE does not characterize the observed data directly but rather the amount of learning performed by a learning model in response to incoming observations. Previous studies demonstrated its applicability to adaptive filters, in-parameter-linear neural architectures, and gradient-based learning systems [16], [17], [19], where elevated LE values were associated with increased information acquisition during learning.

Recently, LE was extended from temporal adaptive systems to image analysis through Spatial Learning Entropy Map (SLEM) generated by a multilayer perceptron (MLP) network [20]. A distinctive property of SLEMs is that they are generated by incremental sample-wise learning rather than by simultaneous processing of the entire image. Local pixel neighborhoods are presented sequentially to a neural predictor according to a predefined traversal path through the image. Consequently, the learning induced by a particular image location depends not only on its local structure but also on the internal representation formed from previously processed locations. The resulting LE values therefore reflect both image structure and the learning trajectory of the neural model. In this sense, SLEMs characterize how image structures are progressively discovered during learning rather than how they appear in the image alone.

Our previous work demonstrated that elevated LE values tend to concentrate around corners, contour intersections, shape singularities, and other structurally significant image locations [20]. However, while SLEMs were shown to identify regions associated with increased learning activity, the discriminative relevance of such regions remains unexplored. In particular, it is not known whether locations corresponding to elevated learning activity also preserve information useful for image classification and recognition.

This paper addresses this question through the introduction of Learning Entropy Signatures (LES), a compact image descriptor derived from the spatial distribution of $K$-largest-LE locations, and the main contributions of this paper are threefold. First, the concept of Learning Entropy is extended from structural point extraction toward image representation and classification. Second, a novel descriptor termed Learning Entropy Signature is introduced. Third, the obtained results provide additional insight into the relationship between incremental, i.e., pixel-wise, learning activity and discriminative information in image data.

The remainder of the paper is organized as follows. Section II recalls recently introduced Spatial Learning Entropy Maps. Section III introduces the newly proposed Learning Entropy Signature. Section IV presents the experiments. Section V discusses the obtained results and relevant aspects. Finally, Section VI concludes the paper.

## II. Spatial Learning Entropy Maps

Consider a grayscale image as follows

$$\mathbf{I}(i,j);\, i = 1,2,\ldots,M,\, j = 1,2\ldots,N \qquad (1)$$

where $\mathbf{I}$ is the matrix of intensity of all pixels, $M$ and $N$ are image dimensions in pixels and $i$, $j$ are the pixel coordinates.

Let's train MLP to predict the intensity of locally centered pixels, so called target pixels, where the target neighborhood is input to MLP (i.e. image margin pixels are not used as target pixels, but can be used as input neighborhood).

In a simplest case, the target pixels, i.e. locally centered pixels, are defined as the cropped image as follows

$$\hat{\mathbf{I}}(i,j);\, i = 1,2,\ldots,M-\nu,\, j = 1,2\ldots,N-\nu, \qquad (2)$$

where $\hat{\mathbf{I}}(i,j)$ is the target pixel intensity and $\nu$ is the neighborhood distance in pixels.

For all the target pixels at (1), local neighborhood is extracted and presented (in a on oriented way) to a multilayer perceptron for sample-wise learning that trains MLP to predict the central pixel value. Of course, the directionality of this incremental learning may affect how MLP learns structure of the image, and in this paper, we consider the directionality of incremental learning to be preserved the same (e.g. row wise from left to right and from top to bottom).

Let define matrix of all MLP neural weights as follows

$$\mathbf{W}(k,\epsilon);k = 1,2\ldots M\cdot N\,;\,\epsilon = 1,2\ldots epochs. \qquad (3)$$

where $\mathbf{W}$ are all neural weights of the MLP network, $k$ is index of learning increment within $\epsilon^{th}$ training epoch, where $k$ is changing with target pixel coordinates, i.e.,

$$k=(i-1)\cdot(N-\nu)+j;i=1\ldots M-\nu,j=1\ldots N-\nu, \qquad (4)$$

i.e. $i$, $j$ are as in (2) and that also determines the directionality of incrementally learning MLP, i.e., pixel-wise learning of MLP. In the following text, we omit epoch indexes unless they are needed for clarity. Thus, the actual learning increment at the $\epsilon$-th training epoch is

$$\Delta\mathbf{W}(k) = \mathbf{W}(k+1) - \mathbf{W}(k). \qquad (5)$$

The simplest estimation of Learning Entropy quantity can be based on the unusually large learning activity, e.g. [16], here newly on the given pixel and its neighborhood that can be calculated as

$$LE(k) = \|\Delta\mathbf{W}(k)\|\;;LE \in \langle 0,\infty \rangle, \qquad (6)$$

where $LE(k) = LE(k,i,j,\epsilon)$ is the estimation of the Learning Entropy for the given pixel $k \leftrightarrow [i,j]$ (4) and its neighborhood after last training epoch $\epsilon$.

It is important to note that we propose calculating LE after the MLP pixel predictor has been initially pretrained on the image until the neural weights no longer show significant convergence, so the truly hard-to-learn points remain exposed.

By dividing LE (6) of each pixel by the maximum LE found in the whole image (after the last epoch of learning) as

$$LE(k)=\frac{\|\Delta\mathbf{W}(k)\|}{\max\{\|\Delta\mathbf{W}(k)\|,\forall k\}}\,;LE\in\langle 0,1\rangle, \qquad (7)$$

we obtain normalized quantification of Learning Entropy for individual pixels in an image for MLP NN. The normalization means that pixels and their neighborhood with $LE = 0$ induced zero learning activity of the MLP predictor, while pixels approaching $LE = 1$ induce the largest need for learning them.

As this procedure throughout the image domain yields the 2-D distribution of Learning Entropy for 2-D images, we introduce the Spatial Learning Entropy Map (SLEM) as follows

$$\mathbf{S}=\mathbf{S}(i,j,\epsilon)=\{LE(i,j)\};i=1,\ldots,M-\nu,j=1,\ldots,N-\nu, \qquad (8)$$

where $\mathbf{S}$ is the SLEM, i.e., a two-dimensional distribution of LE on image $\mathbf{I}$, produced by a spatially fixed, pixel-wise trained MLP predictor after the $\epsilon$-th training epoch. The example of SLEM and $K$-largest-LE points for fundamental objects is shown in Fig. 1

## III. Learning Entropy Signature

The central hypothesis is that image locations requiring substantial learning – specifically, unusual retraining – correspond to areas with elevated discriminative information. If this assumption holds, a relatively small subset of the $K$-largest-LE locations may provide a compact yet informative image representation suitable for classification tasks. Unlike conventional descriptors derived from image geometry, local statistics, contour representations, or learned latent features, the proposed Learning Entropy Signature (LES) is constructed directly from the learning behavior of neural weights. Thus, it reflects both image structure and the learning trajectory induced during the final epoch of sample-wise training, when the weights no longer converge.

Let

$$P = \{p_1, p_2, \ldots, p_K\} \qquad (9)$$

denote the set of image locations corresponding to the $K$-largest-LE values of the Spatial Learning Entropy Map. Thus, the points in (9) are selected and ordered according to decreasing Learning Entropy values; hence, the index

$i = 1,\ldots,K$ denotes the rank of a selected location in this ordered set as follows

$$LE_1 \geq LE_2 \geq \cdots \geq LE_K ; LE_1 = \max(\mathbf{S}), \quad (10)$$

where $LE_1$ belongs to the image pixel with its largest Learning Entropy, i.e., to the maximum value in the Spatial Learning Entropy Map $\mathbf{S}$ (8).

The centroids of the point set are computed as

$$c_x = \frac{1}{K}\sum_{i=1}^{K} x_i \, , \, c_y = \frac{1}{K}\sum_{i=1}^{K} y_i . \quad (11)$$

The radial distance of each point from the centroid is

$$r_i = \sqrt{(x_i - c_x)^2 + (y_i - c_y)^2}. \quad (12)$$

Let the maximum radial distance within the selected point set be defined as

$$R_{\max} = \max_{j=1,\ldots,K} r_j . \quad (13)$$

Distances are then normalized according to

$$\hat{r}_i = \frac{r_i}{R_{\max}}, \qquad i = 1,\ldots,K. \quad (14)$$

The radial form of the Learning Entropy Signature is then defined as

$$LES_r = [\hat{r}_1, LE_1, \hat{r}_2, LE_2, \ldots, \hat{r}_K, LE_K]^T. \quad (15)$$

This representation is compact and invariant to translation and uniform scaling of the selected $K$-largest-LE points. However, this invariance applies to the descriptor construction itself and does not imply invariance of the underlying Spatial Learning Entropy Map (which exceeds the scope of this paper).

Nevertheless, because only the distance from the centroid is retained, the arrangement of the selected high-LE points is not explicitly represented. Consequently, two point configurations with different spatial organization may produce similar radial signatures if their distances from the centroid are comparable. To preserve the full two-dimensional organization of the selected high-LE locations, we also define the centered coordinate components

$$r_{x,i} = x_i - c_x \, , \, r_{y,i} = y_i - c_y . \quad (16)$$

The normalized coordinate components are computed as

$$\hat{r}_{x,i} = \frac{x_i - c_x}{R_{\max}}, \quad \hat{r}_{y,i} = \frac{y_i - c_y}{R_{\max}}, \quad (17)$$

where the same normalization factor $R_{\max}$ is used for all selected points.

The corresponding Cartesian Learning Entropy Signature is defined as

$$LES_{xy} = [\hat{r}_{x,1}, \hat{r}_{y,1}, LE_1, \hat{r}_{x,2}, \hat{r}_{y,2}, LE_2, \ldots, \hat{r}_{x,K}, \hat{r}_{y,K}, LE_K] \quad (18)$$

Compared with $LES_r$, the descriptor $LES_{xy}$ retains the directional distribution of high-LE locations around the centroid. This may increase discriminative capability in classification problems where object orientation and asymmetric spatial organization are relevant. The price for this increased specificity is reduced rotational invariance.

Alternatively, the same spatial information may be expressed in polar coordinates by introducing the angular component

$$\theta_i = \mathrm{atan}2(y_i - c_y, x_i - c_x), \quad (19)$$

which yields the spatial alternative to (18) as follows

$$LES_{r\theta} = [\hat{r}_1, \theta_1, LE_1, \hat{r}_2, \theta_2, LE_2, \ldots, \hat{r}_K, \theta_K, LE_K]^T. \quad (20)$$

Thus, $LES_r$ provides the most compact radial representation, whereas $LES_{xy}$ and $LES_{r\theta}$ preserve a more complete description of the spatial arrangement of learning-relevant image locations.

The descriptor therefore combines image geometry with information about the network learning trajectory induced by the sequential pixel-wise presentation of the image, thereby linking structural image characteristics with neural learning dynamics in a unified image representation.

## IV. Experimental Analysis

To illustrate the proposed methodology, Figs. 1–3 show representative images, their corresponding Spatial Learning Entropy Maps obtained from stepwise (sample-wise) learning of a feedforward MLP predictor, and the resulting $K$-largest-LE locations that form the Learning Entropy Signature used as an image descriptor.

**Tab. 1**: Configuration of the feedforward MLP predictors and incremental learning procedure used for Spatial Learning Entropy Map generation and subsequent Learning Entropy Signature extraction in all experiments.

| **I** **image dataset** | Synthetic shapes Fig.1 | MNIST digits Fig.2 [21] | Olivetti faces Fig.3 [22] |
|---|---|---|---|
| **MLP architecture** | (24)-16-1 (inputs)-ReLU-Sigmoid stochastic gradient descent | | |
| **training** | Incremental (sample-wise) | | |
| $\mathcal{T}$ **traversal strategy** | row-wise left-to-right top-to-bottom | | |
| $\nu$ **neighborhood radius (MLP input)** | square 5x5 neighborhood, $\nu=2$ central pixel excluded | | |
| **training epochs** | $\epsilon$ =15 | | |
| **learning rate** | 0.01 | | |

In all experiments, a feedforward MLP predictor with a single hidden layer of ReLU neurons was used. The network was trained to predict the intensity of locally centered pixels from their surrounding neighborhoods. A neighborhood radius of ($\nu=2$) was used, resulting in 24 input pixels and a single output corresponding to the central pixel intensity. The network was pretrained until weight learning converged, using incremental sample-wise learning with stochastic gradient descent (SGD) and mean squared error loss. Learning Entropy values were computed from the weight increments after the final training epoch. Unless otherwise specified, the *K*-largest-LE locations were selected to construct the corresponding Learning Entropy Signatures.

Three image domains of increasing structural complexity were considered. First, a set of synthetic geometric shapes was used to visualize the relationship between object geometry, SLEM distributions, and the resulting LES descriptors (Fig. 1). Second, representative samples from the MNIST handwritten digit dataset [21] were analyzed to investigate the behavior of LES on digits (Fig. 2). Finally, selected images from the Olivetti faces dataset [22] were used to demonstrate the potential (and weaknesses to be researched further) of the proposed approach to facial image structures (Fig. 3). To assess whether LES preserves information relevant for image differentiation, low-dimensional t-SNE projections of the resulting descriptors were generated for the geometric shapes, MNIST digits, and Olivetti face images (Figs. 4–6).

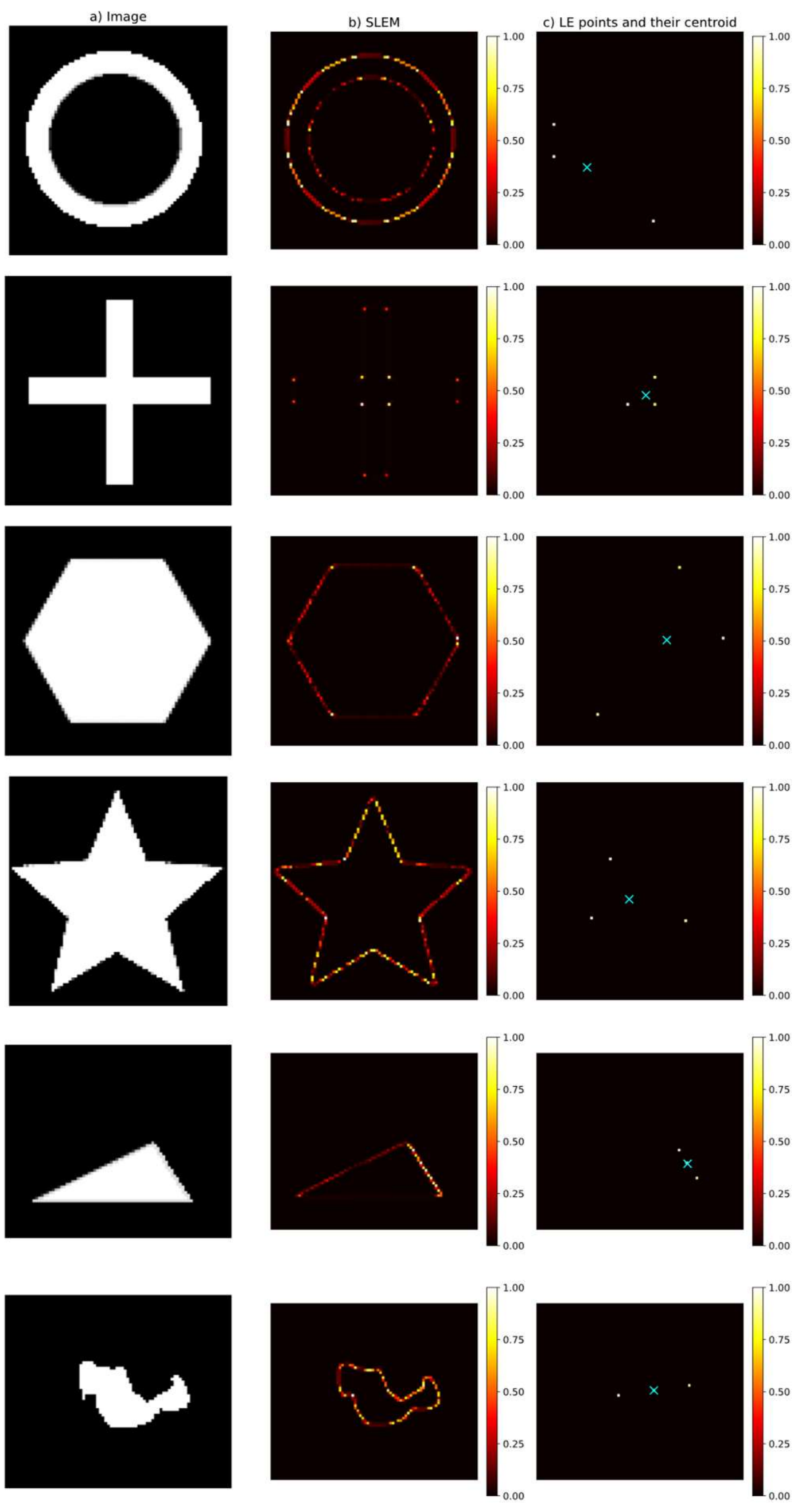

**Fig. 1**: Construction steps for the Spatial LE Map (SLEM) (1)-(8) and Learning Entropy Signature (LES) (9)-(20) on synthetic structures: a) Image, b) SLEM displaying increased learning activity for image structures that are unexpected by MLP during learning (Tab.1), c) *K*-largest-LE points (*K*=3) and the centroid.

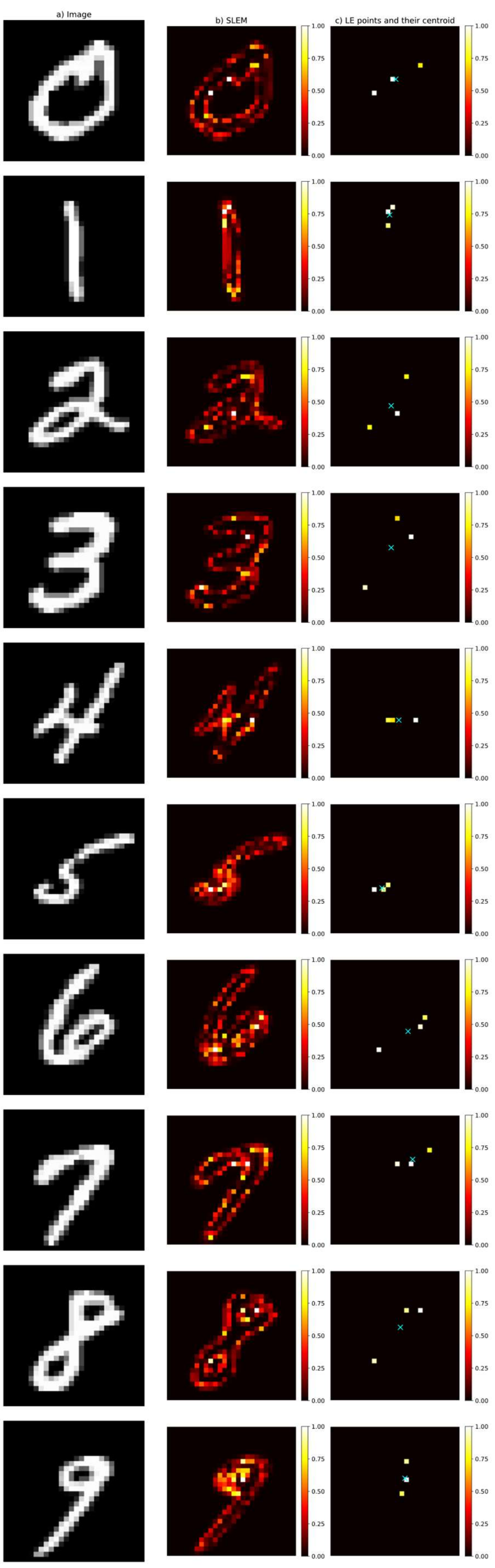

**Fig. 2**: a) Original image, b) SLEM via learning MLP (Tab.1), c) *K*-largest-LE points (*K*=3) and their centroids from the MNIST handwritten digit dataset [21].

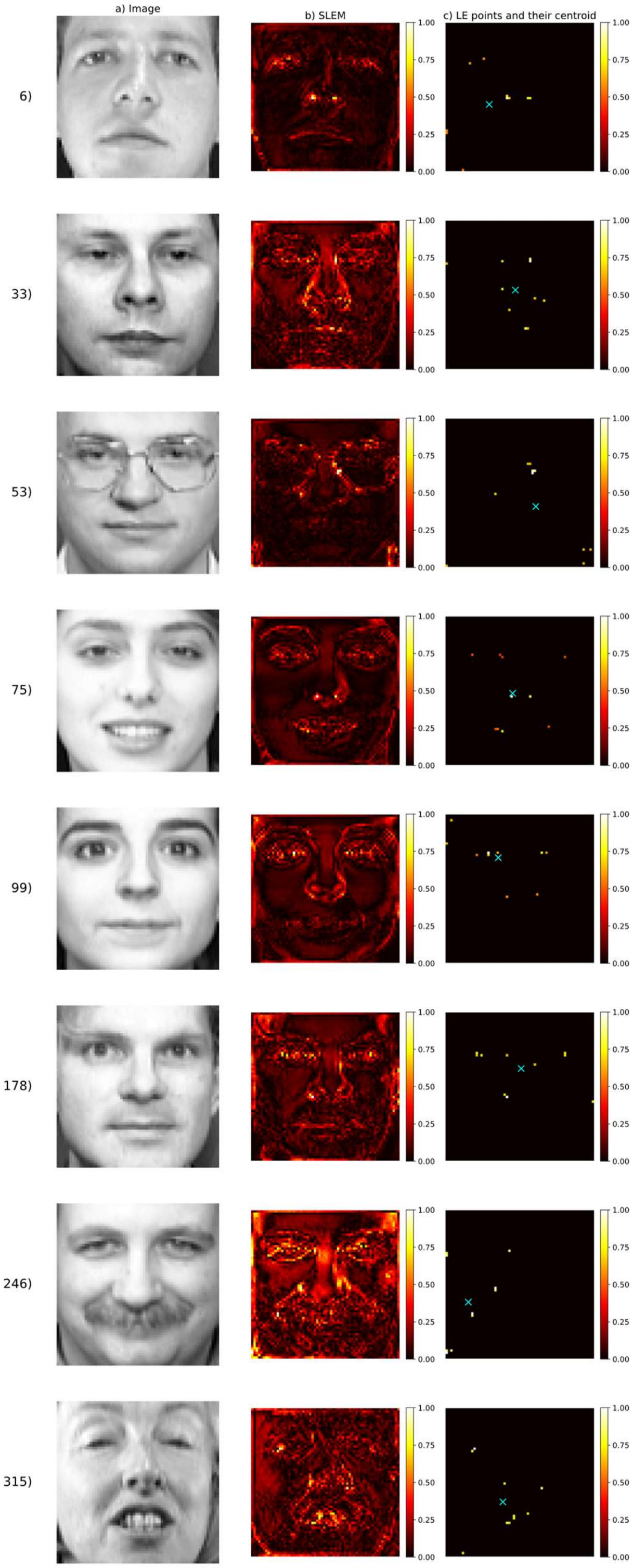

**Fig. 3**: a) Original images from the Olivetti faces dataset [22], b) their SLEM via learning MLP (Tab.1), c) *K*-largest-LE points and their centroids (K=10).

Inspection of Figs. 1–3 shows that the *K*-largest-LE values are not distributed uniformly across the images. Instead, elevated LE locations cluster around structurally distinctive regions, such as corners, contour intersections, high-curvature segments, and other geometrically singular locations. This pattern is consistent across synthetic objects, handwritten digits, and facial images, indicating that the resulting Learning Entropy Signatures capture these features. These signatures were then projected into a two-dimensional space using t-SNE.

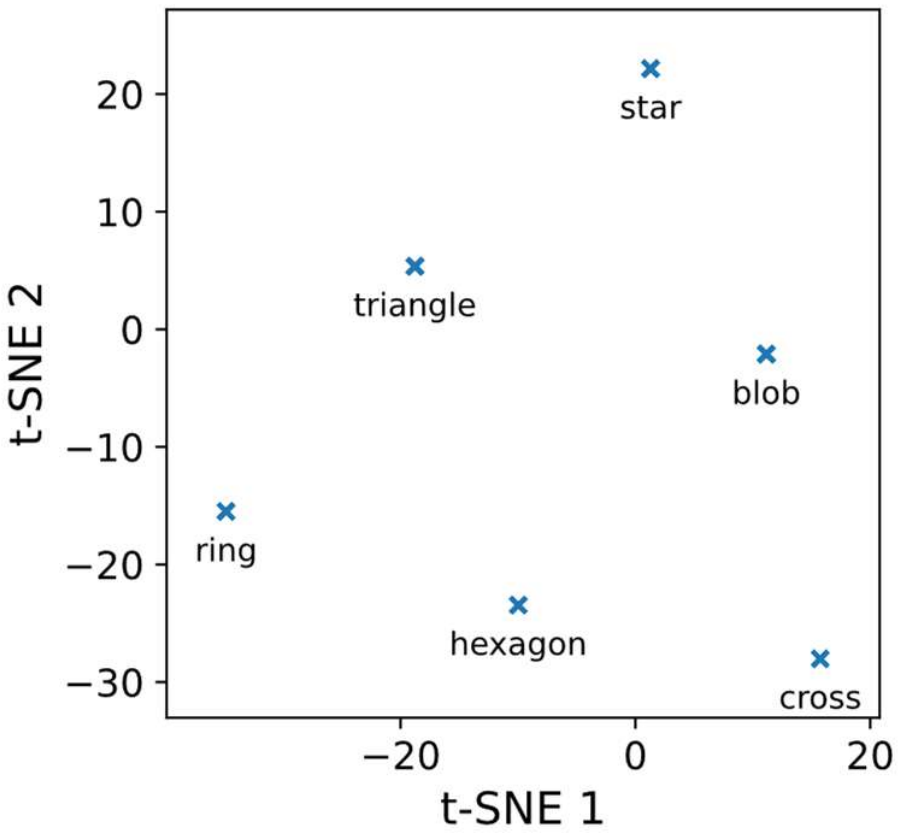

**Fig. 4**: t-SNE 2D projection of the Learning Entropy Signature as image descriptors derived from the geometric shape dataset (Fig. 1). The 2D projection demonstrates the separability of shape classes by the proposed LES.

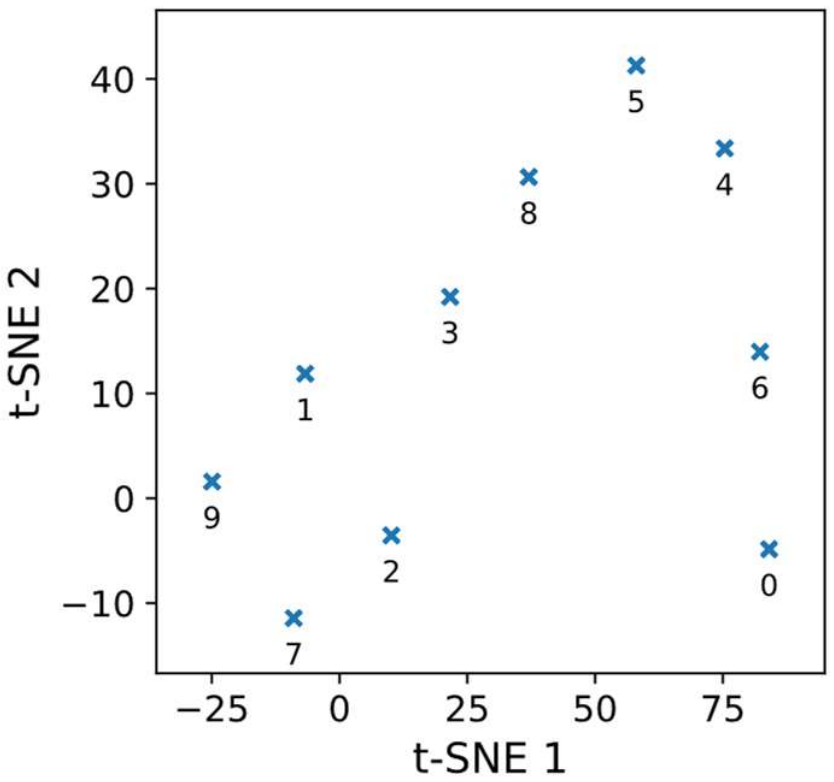

**Fig. 5**: t-SNE 2D projection of the Learning Entropy Signature as image descriptors derived from the handwritten MNIST dataset samples (Fig.2). The 2D projection demonstrates the separability of digit images by the proposed LES.

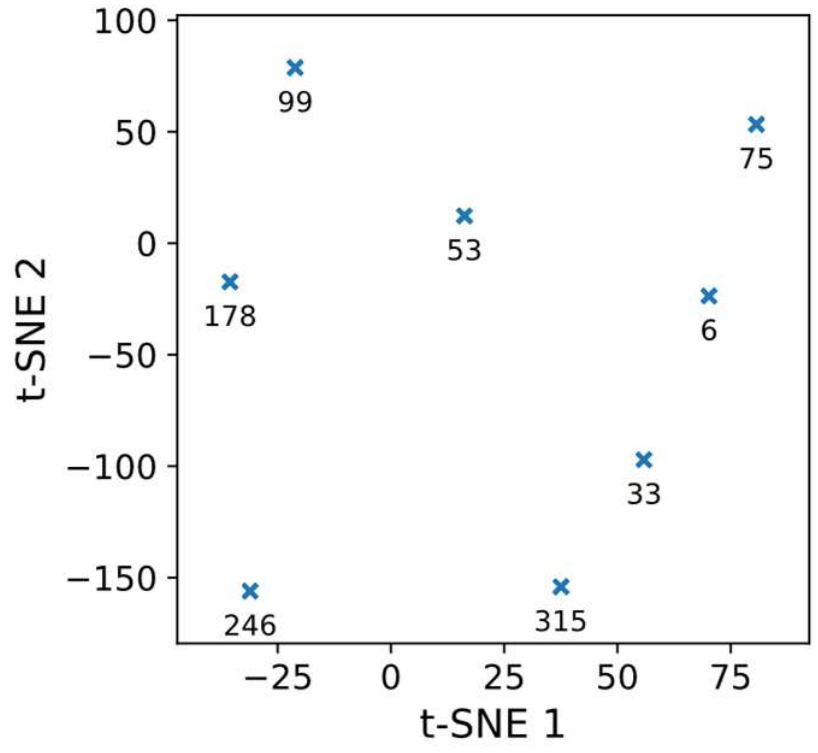

**Fig. 6**: t-SNE 2D projection of the Learning Entropy Signature descriptors derived from the Olivetti faces dataset samples (Fig. 3). The 2D projection demonstrates the separability of images with faces by the proposed LES.

The objective was not classification, but to determine whether the compact descriptors derived solely from *K*-largest-LE locations preserve meaningful differences among images. If the proposed descriptor captures discriminative structural information, images with different structural organization should occupy distinct regions of the projected feature space despite the substantial dimensionality reduction.

## V. Discussion

This study examines whether a compact image descriptor constructed from the $K$-largest-LE points preserves information relevant for distinguishing images. The resulting Spatial Learning Entropy Maps show that elevated LE values consistently appear around structurally distinctive image regions, rather than being uniformly distributed across the image domain. In synthetic shapes, handwritten digits, and face images, high-LE points appear around areas with increased structural complexity. This finding supports interpreting Learning Entropy as a measure reflecting the amount of adaptation required by the learning process, related to object expectancy by a pretrained network on the same image.

A notable result is that only a small subset of $K$-largest-LE points is sufficient to construct compact image descriptors that preserve meaningful differences among images. The t-SNE projections in Figs. 4–6 show that LES descriptors derived from different shapes, digits, and face images occupy distinct regions of the descriptor space.

In contrast to conventional image descriptors, LES is not derived directly from image intensities, contours, gradients, or latent representations learned by deep neural networks. Instead, it is based on the adaptation dynamics of a neural predictor trained incrementally in an oriented, sample-wise manner, i.e, the traversal strategy $\mathcal{T}$ . Consequently, LES is not solely a function of image geometry, that is, $LES = f(\mathbf{I})$ , as is common for conventional descriptors, but rather

$$LES = f(\mathbf{I}, \mathcal{T}, \mathbf{W}_\infty), \tag{21}$$

where $\mathcal{T}$ denotes the traversal strategy used for incremental sample-wise learning on image $\mathbf{I}$ , and $\mathbf{W}_\infty$ represents the neural weights of the converged pretrained MLP predictor (practically obtained during last epoch of training). As a result, the descriptor captures not only structural properties of the image but also the response of the learning system to those structures. Unlike conventional geometric descriptors, LES is not inherently invariant to image transformations, since modifications to the image may alter the learning dynamics and the resulting Spatial Learning Entropy Map. The translation and scaling invariance discussed in Section III applies only to the normalized spatial representation of the selected $K$-largest-LE points. Thus, LES links image representation with the learning trajectory that emerges during sequential image exploration.

The principal contribution of this work is not a new image recognition method, but the introduction of a learning-oriented image descriptor derived from neural adaptation activity. The results support the hypothesis that image structures inducing increased learning activity contain information that remains relevant for image differentiation even after substantial descriptor compression. Future work should investigate this relationship quantitatively through classification experiments, larger image datasets, alternative image traversal strategies, and comparisons with established image descriptors.

## VI. Conclusions

This paper introduces the Learning Entropy Signature (LES), a compact image descriptor derived from the spatial distribution of the $K$-largest-LE points obtained from Spatial Learning Entropy Maps. Unlike conventional image descriptors, LES is constructed from the learning adaptation activity of a feedforward MLP predictor during incremental sample-wise learning, thus reflecting both image structure and the learning trajectory induced by the traversal strategy. Experiments on synthetic shapes, handwritten digits, and face images demonstrate that a relatively small number of high-LE points preserves structural differences among images and yields distinguishable descriptor representations in a low-dimensional feature space. The results suggest a relationship between elevated learning activity and image structures that carry discriminative information. Future work will focus on quantitative classification experiments, larger image datasets, and further investigation of the influence of learning and traversal parameters on the resulting descriptors.

## Acknowledgements

This work was supported in part by the Student Grant Competition (SGS) of Czech Technical University in Prague, project No. SGS26/186/OHK3/3T/18, in part by the Operational Programme Johannes Amos Comenius, project "MEDDA Medical Database (Evolution in Medical Diagnostics)" No. CZ.02.01.01/00/23_021/0012422 of the Ministry of Education, Youth and Sports of the Czech Republic.", and in part by the Grant for Cancer Screening Research from the Japan Cancer Society, by the Subsidy for Support Projects for New Entries and New Industry Creation from Miyagi Prefectural Government, and by the JST COI-NEXT Vision to Connect.